\documentclass[review]{elsarticle}

\journal{Neurocomputing}

\usepackage[numbers]{natbib}
\usepackage{graphicx}
\usepackage{multirow}

\usepackage{caption} 
\usepackage{bm}
\usepackage{amssymb}
\usepackage{amsmath}
\usepackage[ruled]{algorithm2e}
\usepackage{booktabs}
\usepackage{color}

\begin{document}
	
	\begin{frontmatter}
		
		\title{Multi-view Subspace Clustering Networks with Local and Global Graph Information}

		\author[mymainaddress]{Qinghai Zheng}
		
		\author[mymainaddress]{Jihua Zhu}
		\cortext[mycorrespondingauthor]{Corresponding author}
		\ead{zhujh@xjtu.edu.cn}
		
		\author[mymainaddress]{Yuanyuan Ma}
		
		\author[mymainaddress]{Zhongyu Li}
		
		\author[mymainaddress]{Zhiqiang Tian}
		
		\address[mymainaddress]{School of Software Engineering, Xi'an Jiaotong University, Xi'an 710049, China}
		
		\begin{abstract}
			This study investigates the problem of multi-view subspace clustering, the goal of which is to explore the underlying grouping structure of data collected from different fields or measurements. Since data do not always comply with the linear subspace models in many real-world applications, most existing multi-view subspace clustering methods based on the shallow linear subspace models may fail in practice. Furthermore, the underlying graph information of multi-view data is usually ignored in most existing multi-view subspace clustering methods. To address the aforementioned limitations, we proposed the novel multi-view subspace clustering networks with local and global graph information, termed MSCNLG, in this paper. Specifically, autoencoder networks are employed on multiple views to achieve latent smooth representations that are suitable for the linear assumption. Simultaneously, by integrating fused multi-view graph information into self-expressive layers, the proposed MSCNLG obtains the common shared multi-view subspace representation, which can be used to get clustering results by employing the standard spectral clustering algorithm. As an end-to-end trainable framework, the proposed method fully investigates the valuable information of multiple views. Comprehensive experiments on six benchmark datasets validate the effectiveness and superiority of the proposed MSCNLG.
		\end{abstract}
		
		
	\end{frontmatter}
	
	\section{Introduction}
	
	Clustering is an important task in unsupervised learning, which can be a prepossessing step to assist other learning tasks or a stand-alone exploratory tool to uncover underlying information of data points \cite{zhou2012ensemble_methods}. The goal of clustering is to group unlabeled data points into corresponding categories according to their intrinsic similarities. Many effective clustering algorithms have been proposed, such as k-means clustering, spectral clustering \cite{spectral_clustering_tutorial}, and subspace clustering \cite{sparse_subspace_clustering_PAMI2013,LRR_PAMI2012,SmoothRepresentationClustering}. And some deep-learning based clustering methods are also proposed in recent years \cite{DeepSubspaceClustering,LDPDSC}. However, these methods are designed for single-view rather than multi-view data, which are collected from multiple sources and common in many real-world applications. \textcolor{black}{For example, in the task of multimedia content understanding, visual frames and audio signals are two distinct views of a video and both are important for this task. In medical domain, the computerized tomography and magnetic resonance image are two different views and are vital for the diagnosis of laryngeal cyst.} Unlike single-view data, multi-view data contains both the consensus information and complementary information for multi-view learning \cite{multiview_learning_survey}. Therefore, an important issue of multi-view clustering is how to fuse multiple views properly to mine the underlying grouping information effectively. Evidently, it is not a good choice to use a single-view clustering algorithm on multi-view data straightforward \cite{co_reg_spectral,huang2018robustNC,GLMSC_PAMI2018,2020PartitionKangZhao,2019AutoPR2020,2018SelfKBS}. In this study, we consider the multi-view clustering problem based on the subspace clustering algorithm \cite{LRR_PAMI2012,wang2016multi,DeepSubspaceClustering,ASC_IJCAI2019}, which utilizes the linear subspace model for clustering. To be clear, the linear subspace model assumes that a data point can be represented by a linear combination of other points in the same cluster.
	
	Recently, numerous multi-view subspace clustering methods have been proposed \cite{LtMSC,GLMSC_PAMI2018,MLRSSC_PR2018,multiview_subspace_clustering_dual_ZhouTao,FCMSC,zheng2020constrained,kang2020large}. For example, LT-MSC \cite{LtMSC} boosts the clustering performance by employing a low-rank tensor constraint on multiple views, FCMSC \cite{FCMSC} obtains the promising multi-view clustering performance by introducing the concept of cluster-specific corruption. {\color{black}{PMSC \cite{2020PartitionKangZhao} attains multi-view clustering results by introducing a fusion mechanism, which assigns weights to each basic partition and learns the consensus partition.}} Despite good clustering results can be obtained, there are some deficiencies in these methods. Firstly, the aforementioned multi-view subspace clustering methods are all under the linear subspace assumption \cite{sparse_subspace_clustering_PAMI2013}, in other words, the data in different views are all conformed to the linear subspace models. However, correlations of data points are preferred to be non-linear rather than linear for data in practice. Some works use the kernel methods to deal with this issue by mapping samples into a high-dimensional feature space and then get subspace clustering results in that space \cite{patel2014kernel,xiao2015robust}. Whereas, the kernel function is often pre-defined in an ad hoc manner and suffers from the problem of generalization. Secondly, few multi-view subspace clustering methods integrate the graph information of different views into the subspace representation for improving clustering results. Since graph information is vital for clustering, it is highly expected to achieve improvement for multi-view subspace clustering performance by leveraging the underlying graph information of multi-view data.
	
	\begin{figure*}[t]
		\centering
		\includegraphics[width=0.95\textwidth]{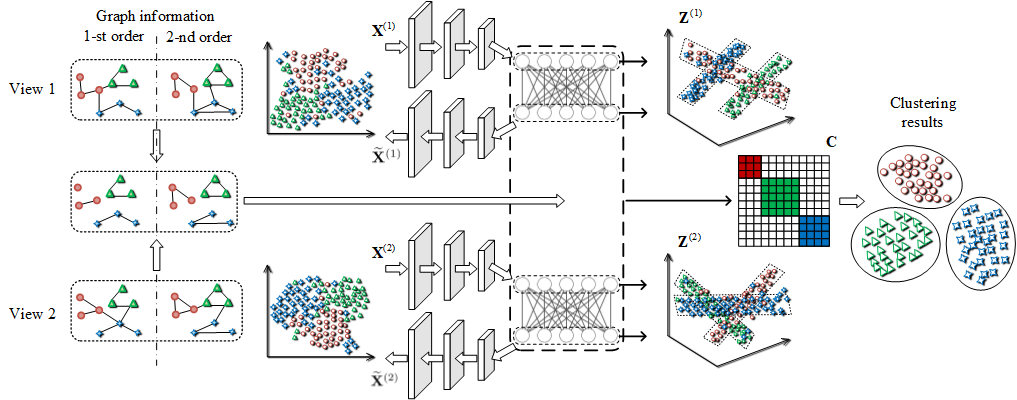} 
		\caption{Illustration of the proposed MSCNLG. The autoencoder networks are performed on different views simultaneously to attain latent representations that are conformed with linear subspace assumption. ${\bm{X}}^{(k)}$ denote the input feature representations, ${{\widetilde {\bm{X}}}^{(k)}}$ and ${\bf{Z}}^{(k)}$ are output of the decoder network and the latent representation with respect to the $i$-th view, respectively, ${\bf{C}}$ indicates the common shared coefficient matrix.	By leveraging the 1-st order and 2-nd order graph of all views, the underlying local and global multi-view graph information can be obtained as well. Consequently, A shared smooth representation, which can be leveraged to get clustering results, is achieved by integrating the underlying graph information into the self-expressive layer of auteoencoder networks.}
		\label{FrameworkC}
	\end{figure*}
	
	To address the above-mentioned limitations, we proposed the multi-view subspace clustering networks with local and global graph information (MSCNLG) in this paper. As shown in Fig.~\ref{FrameworkC}, to conform with the linear subspace assumption, autoencoder networks are simultaneously performed on multiple views respectively, and the self-expressive layer \cite{DeepSubspaceClustering} is employed between the encoder and decoder. The coefficient matrix of self-expressive layer, i.e. the subspace representation, is shared by all views. It is noteworthy that the self-expressive layer leveraged in our method is much different from existing works \cite{DeepSubspaceClustering,2020DeepPengXi,yu2020dcsr,LDPDSC,DMSCN}, such as DSC-Net \cite{DeepSubspaceClustering}, DPSC \cite{LDPDSC}, and DMSCN \cite{DMSCN}. Specifically, the self-expressive layer employed in DSC-Net, DPSC, and DMSCN leverages the ${\ell _1}$ norm or Frobenius norm to constrain the learning process of subspace representation. Actually, the utilization of ${\ell _1}$ norm or Frobenius norm assumes that the subspace representation is sparse \cite{sparse_subspace_clustering_PAMI2013} or dense \cite{ji2014efficient}, and ignores the important graph information of data. Unlike most existing methods, the proposed MSCNLG utilizes the underlying local and global graph information of multi-view data to guide the subspace representation learning process in the self-expressive layer of autoencoders in multiple views. As can be observed in Fig.~\ref{FrameworkC}, local and global graph information can be mined by the first-order and second-order graph, then are leveraged in the shared self-express layer to constrain the learning process of subspace representation. Once the desired subspace representation is obtained, clustering results can be calculated by employing the standard spectral clustering algorithm. Comprehensive experiments conducted on six real-world datasets illustrate the effectiveness of MSCNLG.
	
	The main contributions of this study are as follows: {\color{black}{
			\begin{itemize}
				\item We propose a novel method, termed Multi-view Subspace Clustering Networks with Local and
				Global graph information (MSCNLG), in this paper. By performing autoencoder networks on all views
				simultaneously, latent representations that conform to the linear subspace model can be achieved and
				leveraged in our method for clustering.
				\item We consider the local and global graph information of multi-view data in the proposed method. By introducing the first-order and second-order graph, the local and global graph information can be explored and used to guide the learning process of the desired subspace representation.
				\item We conduct extensive experiments on six real-world benchmark multi-view datasets. Experimental results demonstrate the superiority of our method over several state-of-the-arts.
	\end{itemize}}}
	
	\section{Related Work}
	In recent years, many multi-view clustering approaches have been proposed. Although various multi-view subspace clustering methods based on different theories are conducted, the key issue of them all is one, i.e., achieving promising clustering results by combining multiple views properly and exploring the underlying clustering structures of multi-view data fully.
	
	Based on the way of views combination, most existing multi-view clustering methods can be classified roughly into three groups \cite{multiview_learning_survey,multiview_clustering_survey,kang2019multipleIJCAI}: co-training or co-regularized \cite{co_training_spectral,co_reg_spectral,huang2020auto}, graph based \cite{2019MultiKBS,RMSC,liu2020joint,AMGL,ma2020multiview,GMC,2018AutoPR2018,chen2019auto,li2020bipartite}, and subspace clustering based methods \cite{DiMSC,yin2015multiNeurCom,LtMSC,GLMSC_PAMI2018,MLRSSC_PR2018,FCMSC,zheng2020constrained,kang2020large}. For example, in the first category, \textcolor{black}{the method proposed in \cite{co_training_spectral} searches for the clustering results agreed among multiple views by the idea of co-training}, the approach proposed in \cite{co_reg_spectral} uses the co-regularized strategy to combine graphs information of different views implicitly to achieves multi-view clustering results. For the graph based methods, RMSC \cite{RMSC} pursues a latent transition probability matrix of all views via low rank and sparse decomposition, and then obtains clustering results based on the standard Markov chain. AMGL \cite{AMGL} achieves multi-view clustering results by assigning auto-weighted factors to multiple views. MCGC \cite{MCGC} achieves clustering results by learning a common shared graph of all views with a constrained Laplacian rank constraint. {\color{black}{GFSC \cite{2019MultiKBS} performs graph fusion and spectral clustering simultaneously to get promising multi-view clustering results.}} Regarding subspace clustering based methods, LMSC \cite{LMSC} seeks an underlying representation, which is the origin of all views, and runs the low-rank representation algorithm on the learning latent representation simultaneously. MLRSSC \cite{MLRSSC_PR2018} aims to learn a joint subspace representation and constructs a shared subspace representation with both the low-rank and sparsity constraints. 
	
	Some deep multi-view clustering methods \cite{DCCAE,DMSCN,AE2_CVPR2019,zhang2019cpm,2019MultiIJCAI,ZhangPAMI2021} are also proposed recently. DCCAE \cite{DCCAE} uses two autoencoders and tries to maximize the canonical correlation between two different views. DMSCN \cite{DMSCN} proposes an affinity fusion-based model by employing the self-expressive layer and leverages Frobenius norm to constrain the subspace representation. {\color{black}{$\rm{AE}^{2}$-Nets \cite{AE2_CVPR2019} propose a nested autoencoder framework to learn an intact representation of multi-view data.}} It is worth noting that most existing multi-view subspace clustering methods, including deep-based multi-view subspace clustering, ignore the graph information of multiple views during the subspace representation learning process.
	
	\section{The Proposed Approach}
	\begin{table}[t]
		\centering
		\caption{Main symbols used in this paper.}
		\resizebox{0.73\textwidth}{!}{
			\smallskip\begin{tabular}{l|l}
				\toprule
				Symbol & Meaning \\
				\midrule
				$n$ & The number of samples. \\
				$v$ & The number of views. \\
				$c$ & The number of clusters.\\
				$d_k$ & The dimension of the original $k$-th view.\\
				${\widehat d_k}$ & The dimension of the latent $k$-th view.\\
				${{\bf A}_i}$ & The $i$-th column of matrix ${\bf A}$.\\
				${{\bf X}^{(k)}} \in {\mathbb{R}^{{d_k} \times n}}$ & The data matrix of the $k$-th view.\\
				${{\bf X}_i^{(k)}} \in {\mathbb{R}^{d_k}}$ & The $i$-th data point from the $k$-th view.\\
				${\bf W}^{(k)}$ & The first-order proximity of the $k$-th view.\\
				${\bf{\widehat W}}^{(k)}$ & The second-order proximity of the $k$-th view.\\
				${{\bf Z}^{(k)}} \in {\mathbb{R}^{{{\widehat d_k}} \times n}}$ & The latent representation the $k$-th view.\\
				${\bf C} \in {\mathbb{R}^{n \times n}}$ & The shared smooth representation matrix.\\
				\bottomrule
			\end{tabular}
		}
		\label{table_symbol}
	\end{table}
	
	In this section, we introduce the proposed method in detail. Fig.~\ref{FrameworkC} presents the whole framework of our MSCNLG, Table~\ref{table_symbol} lists the main symbols employed in this paper. Given a multi-view dataset $ {\bf X} = \{ {{\bf X}^{(k)}}\} _{k = 1}^v$, samples of which is collected from $v$ multiple views, we start with the whole objective function of the proposed MSCNLG in this section. To be clear, the objective function is formulated as follows:
	\begin{equation}
		{\cal L} = \sum\limits_{k = 1}^v {({\cal L}_1^{(k)} +  \alpha {\cal L}_2^{(k)})}  +  \beta {{\cal L}_3},
	\end{equation}
	where ${\cal L}_1^{(k)}$ and ${\cal L}_2^{(k)}$ denote the reconstruction loss and the self-representation loss of the $k$-th view respectively, and the Frobenius-norm is employed here. {\color{black}{${\cal L}_3$ stands for a graph regularizer term}}, which employs the multi-view local and global graph information to guide the learning process of multi-view subspace representation. $\alpha$ and  $\beta$ are the trade-off parameters. 
	
	For ${\cal L}_1^{(k)}$ and ${\cal L}_2^{(k)}$, they are utilized to learn the new feature representations, i.e., $\{{\bf{Z}}^{(k)}\}_{k=1}^v$, for multiple views. Specifically, the autoencoder framework with a self-express layer is employed in ${\cal L}_1^{(k)}$ and ${\cal L}_2^{(k)}$, and they have the following formulas:
	\begin{equation}
		\begin{gathered}
			{\cal L}_1^{(k)} = \frac{1}{2}\left\| {{{\bf{X}}^{(k)}} - {{\widetilde {\bf{X}}}^{(k)}}} \right\|_F^2, \hfill \\
			{\cal L}_2^{(k)} = \frac{1}{2}\left\| {{{\bf{Z}}^{(k)}} - {{\bf{Z}}^{(k)}}{\bf{C}}} \right\|_F^2, \hfill \\ 
		\end{gathered}
	\end{equation}
	where ${{\widetilde {\bf{X}}}^{(k)}}$ and ${\bf{Z}}^{(k)}$ are output of the decoder network and the latent representation with respect to the $k$-th view, respectively. ${\bf{C}}$ indicates the common shared coefficient matrix. Obviously, ${\cal L}_1^{(k)}$ promises that the information of the $k$-th view can be reserved in maximum to learn the latent new representation ${\bf{Z}}^{(k)}$. For ${\cal L}_2^{(k)}$, it leverages $\{{\bf{Z}}^{(k)}\}_{k=1}^v$ and aims to learn the desired subspace representation ${\bf{C}}$ for clustering. It is clear that a specific constraint should be added to guide the learning process of ${\bf{C}}$. In the proposed MSCNLG, we leverage the underlying local and global information of multi-view data by introducing ${\cal L}_3$ to achieve this goal. It is emphasized that ${\cal L}_3$ employed here is much different from existing deep based subspace clustering, which employs the ${\ell _1}$ norm or Frobenius norm to regularize the subspace representation learning process and graph information is neglected.
	
	\begin{figure}[t]
		\centering
		\includegraphics[width=0.65\textwidth]{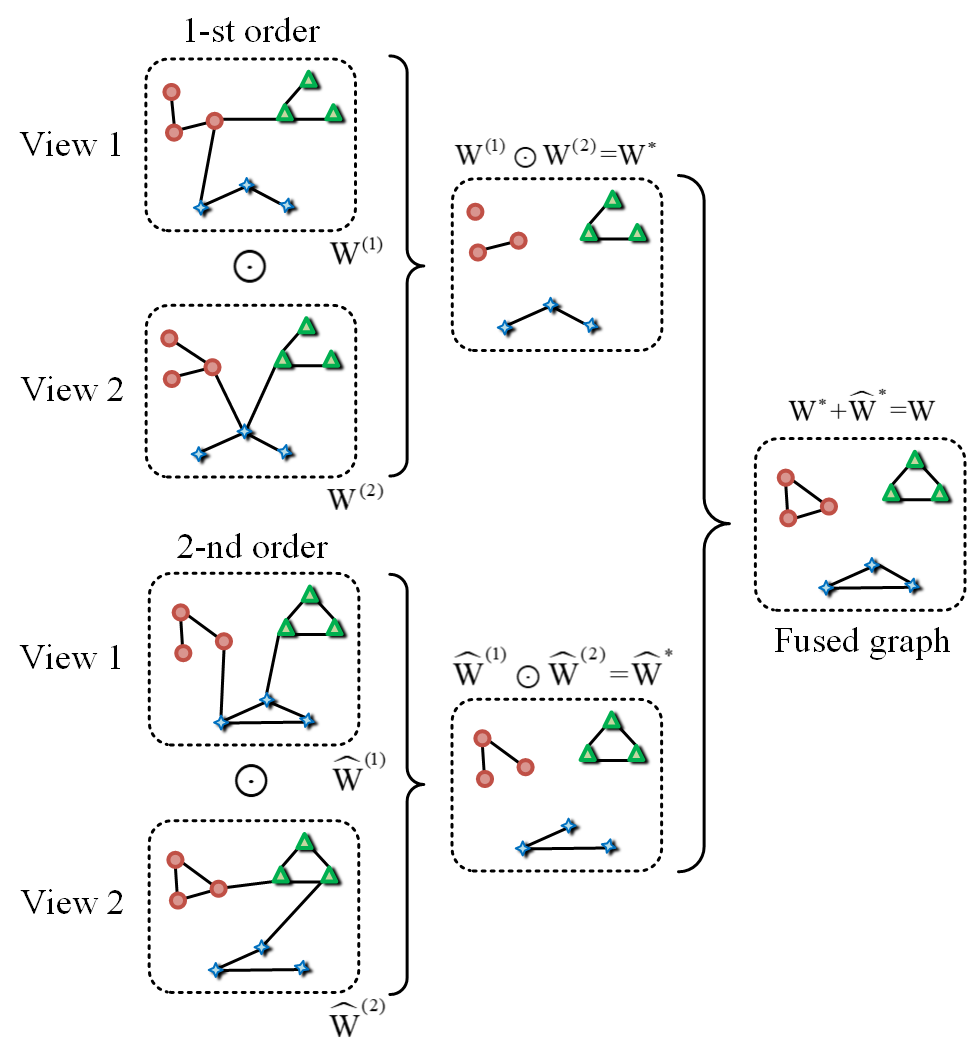} 
		\caption{Illustration of the proposed multi-view graph fusion. $ \odot $ denotes the Hadamard product.}
		\label{Graph_Fusion}
	\end{figure}
	
	Regarding to ${\cal L}_3$, it investigates the underlying local and global graph information of multi-view data and guides the learning process of desired subspace representation. Graph information of data is vital for clustering, however, it is a challenging to explore graph information of multiple views effectively. Here, the first-order proximity and second-order proximity \cite{Graph12Order} of different views are employed to attain a fused graph. {\color{black}{Taking the $k$-th view for example, the similarity between the $i$-th and $j$-th data points is $ \exp (-\frac{{\left\| {{{\bf X}_i ^{(k)}} - {{\bf X}_j ^{(k)}}} \right\|_2^2}}{{{\sigma ^2}}})$, where $\sigma $ denotes the median Euclidean distance. Mutual $k$ nearest neighbor (m$k$NN) strategy is employed in our method, which means that the elements of the first-order proximity are:
			\begin{equation}
				{{\bf W} ^{(k)}} = \left\{ \begin{array}{l}
					\exp (-\frac{{\left\| {{{\bf X}_i ^{(k)}} - {{\bf X}_j ^{(k)}}} \right\|_2^2}}{{{\sigma ^2}}}),{\kern 1pt} {\kern 1pt} {\kern 1pt} {\rm{if}}{\kern 1pt} {\kern 1pt} {\bf X}_j^{(v)}{\kern 1pt} {\rm{and}}{\kern 1pt} {\kern 1pt} {\kern 1pt} {\kern 1pt} {\bf X}_i^{(v)}{\kern 1pt} {\rm{are}}{\kern 1pt} {\kern 1pt} {\kern 1pt} {\rm{m\-}}k{\rm{NN}}{\kern 1pt} {\kern 1pt}, \\
					0{\kern 1pt} {\kern 1pt} {\kern 1pt} {\kern 1pt} {\kern 1pt} {\kern 1pt} {\kern 1pt} {\kern 1pt} {\kern 1pt} {\kern 1pt} {\kern 1pt} {\kern 1pt} {\kern 1pt} ,{\kern 1pt} {\kern 1pt} {\rm{otherwise}}
				\end{array} \right.
			\end{equation}
			where ${\bf W}^{(k)}$ is the first-order proximity matrix of the $k$-th view.}} Clearly, ${\bf W}^{(k)}$ captures the local graph structures, and it is oversimplified to fully capture the graph information for clustering. {\color{black}{To this end, the second-order proximity of the $i$-th and $j$-th data points can be formulated as follows:
			\begin{equation}
				{{\bf \widehat W} ^{(k)}} = \left\{ \begin{array}{l}
					\exp (-\frac{{\left\| {{{\bf W}_i ^{(k)}} - {{\bf W}_j ^{(k)}}} \right\|_2^2}}{{{\sigma ^2}}}),{\kern 1pt} {\kern 1pt} {\kern 1pt} {\rm{if}}{\kern 1pt} {\kern 1pt} {\bf W}_j^{(v)}{\kern 1pt} {\rm{and}}{\kern 1pt} {\kern 1pt} {\kern 1pt} {\kern 1pt} {\bf W}_i^{(v)}{\kern 1pt} {\rm{are}}{\kern 1pt} {\kern 1pt} {\kern 1pt} {\rm{m\-}}k{\rm{NN}}{\kern 1pt} {\kern 1pt}, \\
					0{\kern 1pt} {\kern 1pt} {\kern 1pt} {\kern 1pt} {\kern 1pt} {\kern 1pt} {\kern 1pt} {\kern 1pt} {\kern 1pt} {\kern 1pt} {\kern 1pt} {\kern 1pt} {\kern 1pt} ,{\kern 1pt} {\kern 1pt} {\rm{otherwise}}
				\end{array} \right.
			\end{equation}
			Under the intuition that data points with more shared neighbors are more likely to be similar, the second order proximity is determined by the number of common neighbors shared by these two data points.}} It is evident that the second-order proximity matrices of multiple views capture the global graph information.
	
	To integrate local and global graph information, we construct the intrinsic multi-view graph as follows:
	\begin{equation}
		{\bf{W}} = \mathop  \odot \limits_{k = 1}^v {{\bf{W}}^{(k)}} + \mathop  \odot \limits_{k = 1}^v {\widehat {\bf{W}}^{(k)}},
	\end{equation}
	where $ \odot $ denotes the Hadamard product, ${\bf{W}}$ contains the local and global graph information of multi-view data. In ${\cal L}_3$, the learning process of desired subspace representation is regularized as follows:
	\begin{equation}
		\begin{gathered}
			{{\cal L}_3} = \frac{1}{2}\sum\limits_{i,j} {{{\bf{W}}_{ij}}\left\| {{{\bf{C}}_i} - {{\bf{C}}_j}} \right\|_2^2} = Tr({{\bf{C}}^T}{\bf{L}}{\bf{C}}), \hfill \\ 
		\end{gathered}
		\label{loss_smoothness}
	\end{equation}
	where ${\bf{L}}$ denotes the Laplacian matrix of ${{\bf W}}$. \textcolor{black}{For the mean of $Tr(\cdot)$, given a matrix $\bf{A} \in {\mathbb{R}^{n \times n}}$, it is defined as 	$Tr({\bf{A}}) = \sum\limits_{i = 1}^n {{{\bf{A}}_{ii}}}$.}
	
	Therefore, the objective function of the proposed MSCNLG can be rewritten as follows:
	\begin{equation}
		\mathcal{L}{\text{ = }}\frac{1}{2}\sum\limits_{k = 1}^v {(\left\| {{{\bf{X}}^{(k)}} - {{\widetilde {\bf{X}}}^{(k)}}} \right\|_F^2 + \alpha \left\| {{{\bf{Z}}^{(k)}} - {{\bf{Z}}^{(k)}}{\bf{C}}} \right\|_F^2)}  + \beta Tr({{\bf{C}}^T}{\bf{L}}{\bf{C}}).
		\label{Objective_Function}
	\end{equation}
	
	As an end-to-end framework, the proposed MSCNLG can be trained effectively based on Adam \cite{Adam} with learning rate fixed to $1 \times {10^{ - 3}}$. {\color{black}{To be clear, some optimization details of the above objective function are provided as well. Specifically, we rewrite Eq.~(\ref{Objective_Function}) as follows:}
		
		\begin{equation}
			\begin{gathered}
				\mathcal{L}{\text{ = }}\frac{1}{2}\sum\limits_{k = 1}^v {(\left\| {{{\bf{X}}^{(k)}} - {{\widetilde {\bf{X}}}^{(k)}}} \right\|_F^2 + \alpha \left\| {{{\bf{Z}}^{(k)}} - {{\bf{Z}}^{(k)}}{\bf{C}}} \right\|_F^2)}  + \beta Tr({{\bf{C}}^T}{\bf{L}}{\bf{C}}) \hfill \\
				~~= \frac{1}{2}\sum\limits_{k = 1}^v {\sum\limits_{i = 1}^n {({{\left\| {{\bf{x}}_i^{(k)} - {\bf{f}}_i^{(M,k)}} \right\|}^2} + \alpha {{\left\| {{\bf{f}}_i^{({M \mathord{\left/
												{\vphantom {M 2}} \right.
												\kern-\nulldelimiterspace} 2},k)} - {\bf{f}}_i^{({M \mathord{\left/
												{\vphantom {M 2}} \right.
												\kern-\nulldelimiterspace} 2},k)}{\bf{C}}} \right\|}^2})} }  + \beta Tr({{\bf{C}}^T}{\bf{L}}{\bf{C}}), \hfill \\ 
			\end{gathered} 
		\end{equation}
		where ${\bf{f}}_i^{(m,k)}$ denotes the output of the $m$-th layer of autoencoder of the $i$-th sample in the $k$-th view. We have ${\widetilde {\bf{X}}^{(k)}} = [{\bf{f}}_1^{(M,k)},{\bf{f}}_2^{(M,k)}, \cdots ,{\bf{f}}_1^{(M,k)}]$ and ${{\bf{Z}}^{(k)}} = [{\bf{f}}_i^{({M \mathord{\left/
					{\vphantom {M 2}} \right.
					\kern-\nulldelimiterspace} 2},k)},{\bf{f}}_2^{({M \mathord{\left/
					{\vphantom {M 2}} \right.
					\kern-\nulldelimiterspace} 2},k)}, \cdots ,{\bf{f}}_n^{({M \mathord{\left/
					{\vphantom {M 2}} \right.
					\kern-\nulldelimiterspace} 2},k)}]$, in which $M$ is the total number of layers. For ${\bf{f}}_i^{(m,k)}$, we have
		\begin{equation}
			{\bf{f}}_i^{(m,k)} = g({{\bf{W}}^{(m,k)}}{\bf{f}}_i^{(m - 1,k)} + {{\bf{b}}^{(m,k)}}),
		\end{equation}
		where $g(\cdot)$ is the activation function used in the autoencoder networks, and ${\bf{\theta} ^{(k)}} = \{ {{\bf{W}}^{(m,k)}},{{\bf{b}}^{(m,k)}}\} _{m = 1}^M$ denotes the parameters of the autoencoder network of the $k$-th view.
		
		For a clear optimization, we give the  gradient of $\mathcal{L}$ with respect to ${{\bf{W}}^{(m,k)}}$, ${{\bf{b}}^{(m,k)}}$, and $\bf{C}$ as follows:
		\begin{equation}
			\begin{gathered}
				\frac{{\partial \mathcal{L}}}{{\partial {{\bf{W}}^{(m,k)}}}} = \sum\limits_{i = 1}^n {(\bm{\Delta} _i^{(m,k)} + \alpha \bm{\Lambda} _i^{(m,k)}){{({\bf{f}}_i^{(m - 1,k)})}^T}} , \hfill \\
				\frac{{\partial \mathcal{L}}}{{\partial {{\bf{b}}^{(m,k)}}}} = \sum\limits_{i = 1}^n {{\bf{\Delta}} _i^{(m,k)} + \alpha {\bf{\Lambda}} _i^{(m,k)}} , \hfill \\
				\frac{{\partial \mathcal{L}}}{{\partial {\bf{C}}}} = \alpha ({{\bf{Z}}^{{{(k)}^T}}}{{\bf{Z}}^{(k)}}{\bf{C}} - {{\bf{Z}}^{{{(k)}^T}}}{\bf{Z}^{(k)}}) + \beta ({{\bm{L}}^T} + {\bf{L}}){\bf{C}}, \hfill \\ 
			\end{gathered} 
		\end{equation}
		where ${\bf{\Delta}} _i^{(m,k)}$ and ${\bf{\Lambda}} _i^{(m,k)}$ have the following definitions:
		\begin{equation}
			{\bf{\Delta}} _i^{(m,k)} = \left\{ \begin{gathered}
				{- ({\bf{x}}_i^{(k)} - {\bf{f}}_i^{(m,k)}) \odot g'({\bf{y}}_i^{(m,k)})},m = M \hfill \\
				{{({{\bf{W}}^{(m + 1,k)}})^T}{\bf{\Delta}} _i^{(m + 1,k)} \odot g'({\bf{y}}_i^{(m,k)})},m < M \hfill \\ 
			\end{gathered}  \right.
		\end{equation}
		and
		\begin{equation}
			{\bf{\Lambda}} _i^{(m,k)} = \left\{ \begin{gathered}
				0,m \geqslant \frac{M}{2} + 1 \hfill \\
				{({\bf{f}}_i^{(m,k)} - {{\bf{c}}_i} - f_i^{(m,k)}{\bf{c}_i}^T{{\bf{c}}_i}) \odot g'({\bf{y}}_i^{(m,k)})},m = \frac{M}{2} \hfill \\
				{{({{\bf{W}}^{(m + 1,k)}})^T}{\bf{\Lambda}} _i^{(m + 1,k)} \odot g'({\bf{y}}_i^{(m,k)})},m \leqslant \frac{M}{2} - 1 \hfill \\ 
			\end{gathered}  \right.
		\end{equation}
		where ${\bf{c}}_i$ denotes the $i$-th column of $\bf{C}$, and ${\bf{y}}_i^{(m,k)} = {{\bf{W}}^{(m,k)}}{\bf{f}}_i^{(m - 1,k)} + {{\bf{b}}^{(m,k)}}$.}
	
	Once the deisred subspace representation ${{\bf C}}$ is optimized, {\color{black}{we can get multi-view clustering results by running the standard spectral clustering algorithm on $\frac{1}{2}(\left| {{\bf C}} \right| + \left| {{{{\bf C}}^T}} \right|)$}}.

	\subsection{Computational complexity}
	{\color{black}{
	For	clarification, we denote $n$, $v$, and $d_{\rm{max}}$ as the number of samples, views, and the maximum dimensionality of multiple views. The main computational burden of the proposed MSCNLG is composed of three parts, including a) construction of graphs, b) optimization of neural networks, and c) spectral algorithm conducted on the learned subspace representation.
			
			For the first part, i.e., construction of graphs, the complexity can be achieved as  $\mathcal{O}(vd_{\rm{max}}n^2)$. For the optimization of neural networks, the complexity is $\mathcal{O}(vn)$. Regarding to the last part, it is clear that the complexity is $\mathcal{O}(n^3)$. Consequently, the total computational complexity of the proposed MSCNLG is $\mathcal{O}(vd_{\rm{max}}n^2+vn+n^3)$.}}

	\section{Experiments}
	
	\begin{table}[t]
		\caption{Statistic information of six multi-view datasets.}
		\label{Datasets_Statistic}
		\centering
		\resizebox{0.5\textwidth}{!}{
			\begin{tabular}{l|c|c|c}
				\toprule
				Dataset & \# Samples & \# Clusters & \# Views\\
				\midrule
				Yale Face & 165 & 15 & 3  \\
				\midrule
				ORL & 400 & 40 & 3  \\
				\midrule
				MSRCV1 & 210 & 7 & 6  \\
				\midrule
				BBC & 685 & 5 & 3  \\
				\midrule
				Caltech101-20 & 2386 & 20 & 6  \\
				\midrule
				LandUse-21 & 2100 & 21 & 3  \\
				\bottomrule
			\end{tabular}
		}
	\end{table}
	
	Comprehensive experiments are performed on six benchmark datasets in this section. To be specific, Yale Face\footnote{http://cvc.yale.edu/projects/yalefaces/yalefaces.html}, which contains 165 images of 15 individuals, and ORL\footnote{https://www.cl.cam.ac.uk/research/dtg/}, which contains 400 images of 40 individuals, are both face images datasets, and each image is described by three features, namely intensity, LBP, and Gabor. MSRCV1\footnote{http://research.microsoft.com/en-us/project/} consists of 210 image samples collected from 7 clusters with 6 views, including CENT, CMT, GIST, HOG, LBP, and SIFT. {\color{black}{BBC\footnote{http://mlg.ucd.ie/datasets/bbc.html}}} consists of 685 new documents from BBC and each of which is divided into 4 sub-parts. Caltech101-20\footnote{http://www.vision.caltech.edu/Image Datasets/Caltech101/} consists of 2386 samples collected from 20 clusters with 6 different features, i.e. Gabor, WM, CENT, HOG, GIST, and LBP. LandUse-21\footnote{http://weegee.vision.ucmerced.edu/datasets/landuse.html} has 2100 samples collected from 21 categories, and it has three types of feature, including GIST, PHOG, and LBP. We summary the statistic information of these multi-view datasets in Table~\ref{Datasets_Statistic}. Four evaluation metrics \cite{LMSC} are utilized, including Normalized Mutual Information (NMI), ACCuracy (ACC), F-measure, and Rand Index (RI). Higher values of all metrics indicate better clustering performance. {\color{black}{Besides, the average rank (Avg. Rank) of these metrics are also reported in this section.}} 
	
	\subsection{Experimental Settings}
	
	\begin{table}[!tbp]
		\centering
		\caption{Clustering results of the comparison experiments on Yale Face, ORL, and MSRCV1. {\color{black}{For NMI, ACC, F-measure, and RI, results in the form of ``mean(std)'' are reported here.}} Values in the bold type denote the best clustering results.}
		\resizebox{1\textwidth}{!}{
			\begin{tabular}{l|l|c|c|c|c|c}
				\toprule
				&Method & NMI& ACC& F-measure& RI & Avg. Rank\\
				\midrule
				\multirow{10}{*}{Yale}  & AMGL  & 0.6437 (0.0192) & 0.6046 (0.0399) & 0.3986 (0.0323)  & 0.9087 (0.0130) & 9.25 \\
				& LMSC        & 0.7011 (0.0096) & 0.6691 (0.0095) & 0.5031 (0.0151)  & 0.9337 (0.0026) & 5.50 \\
				& MLRSSC      & 0.7005 (0.0311) & 0.6733 (0.0384) & 0.5399 (0.0377)  & 0.9420 (0.0049) & 4.50 \\
				& DMSCN       & 0.6889 (0.0121) & 0.7091 (0.0100) & 0.5859 (0.0127)  & 0.9392 (0.0023) & 4.50 \\
				& GMC         & 0.6892 (0.0000) & 0.6545 (0.0000) & 0.4801 (0.0000)  & 0.9257 (0.0000) & 7.00 \\
				& \textcolor{black}{PMSC}  & \textcolor{black}{0.6412 (0.0285)} & \textcolor{black}{0.6146 (0.0457)} & \textcolor{black}{0.4557 (0.0292)} & \textcolor{black}{0.9271 (0.0046)} & \textcolor{black}{8.00} \\
				& \textcolor{black}{mPAC} & \textcolor{black}{0.7227 (0.0000)} & \textcolor{black}{0.6788 (0.0000)} & \textcolor{black}{0.5576 (0.0000)}  & \textcolor{black}{0.9412 (0.0000)} & \textcolor{black}{3.75} \\
				& \textcolor{black}{COMIC} & \textcolor{black}{0.6275 (0.0182)} & \textcolor{black}{0.5352 (0.0328)} & \textcolor{black}{0.4180 (0.0308)}  & \textcolor{black}{0.9146 (0.0104)} & \textcolor{black}{9.50}\\
				\cmidrule{2-7}
				& ${\text{MSCNL}}{{\text{G}}_{{\text{1-st}}}}$& 0.8712 (0.0032) & 0.8910 (0.0033) & 0.8199 (0.0033)  & 0.9738 (0.0007) & 2.00 \\
				& MSCNLG      & \bf{0.9012 (0.0013)} & \bf{0.9152 (0.0000)} & \bf{0.8551 (0.0020)}  & \bf{0.9795 (0.0002)} & 1.00 \\
				\midrule
				\multirow{10}{*}{ORL}    & AMGL        & 0.8830 (0.0149) & 0.6046 (0.0388) & 0.3986 (0.0777)  & 0.9087 (0.0102) & 9.00 \\
				& LMSC        & 0.9215 (0.0168) & 0.8193 (0.0359) & 0.7623 (0.0419)  & 0.9884 (0.0022) & 4.00 \\
				& MLRSSC      & 0.9102 (0.0113) & 0.8042 (0.0233) & 0.7459 (0.0281)  & 0.9879 (0.0014) & 5.00 \\
				& DMSCN       & 0.9532 (0.0004) & 0.8658 (0.0035) & 0.8885 (0.0011)  & 0.9926 (0.0001) & 2.75 \\
				& GMC         & 0.8571 (0.0000) & 0.6325 (0.0000) & 0.3599 (0.0000)  & 0.9357 (0.0000) & 9.25 \\
				& \textcolor{black}{PMSC}        & \textcolor{black}{0.8338 (0.0144)} & \textcolor{black}{0.6688 (0.0260)} & \textcolor{black}{0.5493 (0.0358)}  & \textcolor{black}{0.9738 (0.0032)} & \textcolor{black}{8.50} \\
				& \textcolor{black}{mPAC} & \textcolor{black}{0.8672 (0.0000)} & \textcolor{black}{0.6750 (0.0000)} & \textcolor{black}{0.6282 (0.0000)}  & \textcolor{black}{0.9791 (0.0000)} & \textcolor{black}{7.25} \\
				& \textcolor{black}{COMIC} & \textcolor{black}{0.8854 (0.0090)} & \textcolor{black}{0.7274 (0.0236)} & \textcolor{black}{0.6363 (0.0382)}  & \textcolor{black}{0.9802 (0.0032)} & \textcolor{black}{6.00} \\
				\cmidrule{2-7}
				&${\text{MSCNL}}{{\text{G}}_{{\text{1-st}}}}$& 0.9538 (0.0019) & 0.8700 (0.0058) & 0.8873 (0.0048)  & 0.9929 (0.0003) & 2.25 \\
				& MSCNLG      & \bf{0.9561 (0.0020)} & \bf{0.8975 (0.0029)} & \bf{0.9000 (0.0036)}  & \bf{0.9939 (0.0002)} & 1.00 \\
				\midrule
				\multirow{10}{*}{MSRCV1} & AMGL        & 0.7357 (0.0281) & 0.7171 (0.0847) & 0.6445 (0.0601)  & 0.8807 (0.0365) & 7.75 \\
				& LMSC        & 0.6149 (0.0622) & 0.6948 (0.0734) & 0.5909 (0.0699)  & 0.8826 (0.0205) & 9.75 \\
				& MLRSSC      & 0.6709 (0.0352) & 0.7775 (0.0497) & 0.6524 (0.0480)  & 0.9025 (0.0134) & 7.25 \\
				& DMSCN       & 0.6999 (0.0030) & 0.8000 (0.0002) & 0.6951 (0.0032)  & 0.9028 (0.0014) & 6.25    \\
				& GMC         & 0.8200 (0.0000) & 0.8952 (0.0000) & 0.7997 (0.0000)  & 0.9434 (0.0000) & 2.75 \\
				& \textcolor{black}{PMSC } & \textcolor{black}{0.6477 (0.0448)} & \textcolor{black}{0.7086 (0.0838)} & \textcolor{black}{0.6149 (0.0537)}  & \textcolor{black}{0.8853 (0.0197)} & \textcolor{black}{8.75} \\
				& \textcolor{black}{mPAC}  & \textcolor{black}{0.7204 (0.0000)} & \textcolor{black}{0.8095 (0.0000)} & \textcolor{black}{0.7171 (0.0000)}  & \textcolor{black}{0.9167 (0.0000)} & \textcolor{black}{4.75}\\
				& \textcolor{black}{COMIC}  & \textcolor{black}{0.7362 (0.0229)} & \textcolor{black}{0.8170 (0.0290)} & \textcolor{black}{0.6963 (0.0281)}  & \textcolor{black}{0.9106 (0.0099)} & \textcolor{black}{4.50} \\
				\cmidrule{2-7}
				& ${\text{MSCNL}}{{\text{G}}_{{\text{1-st}}}}$ & 0.8192 (0.0000) & 0.9000 (0.0000) & 0.8254 (0.0000)  & 0.9463 (0.0000) & 2.25\\
				& MSCNLG      & \bf{0.8724 (0.0040)} & \bf{0.9381 (0.0026)} & \bf{0.8867 (0.0043)}  & \bf{0.9666 (0.0012)} & 1.00\\
				\bottomrule
			\end{tabular}
		}
		\label{Comparision_Expreimental_Results}
	\end{table}
	
	\begin{table}[!tbp]
		\centering
		\caption{Clustering results of the comparison experiments on BBC, Caltech101-20, and LandUse-21. {\color{black}{For NMI, ACC, F-measure, and RI, results in the form of ``mean(std)'' are reported here.}} Values in the bold type denote the best clustering results.}
		\resizebox{1\textwidth}{!}{
			\begin{tabular}{l|l|c|c|c|c|c}
				\toprule
				&Method & NMI& ACC& F-measure& RI & Avg. Rank\\
				\midrule
				\multirow{10}{*}{BBC} & AMGL  & 0.5185 (0.0725) & 0.6261 (0.0698) & 0.6050 (0.0527) & 0.7370 (0.0791) & 7.25 \\
				& LMSC        & 0.5594 (0.0409) & 0.7394 (0.0671) & 0.6291 (0.0491) & 0.8336 (0.0223) &5.50 \\
				& MLRSSC      & 0.6935 (0.0004) & 0.8556 (0.0004) & 0.7897 (0.0003) & 0.9022 (0.0002) & 2.00 \\
				& DMSCN       & 0.5773 (0.0038) & 0.7796 (0.0020) & 0.6602 (0.0032) & 0.8425 (0.0014) & 4.00\\
				& GMC         & 0.5628 (0.0000) & 0.6934 (0.0000) & 0.6333 (0.0000) & 0.7664 (0.0000) & 5.50 \\
				& \textcolor{black}{PMSC} & \textcolor{black}{0.0087 (0.0000)} & \textcolor{black}{0.3229 (0.0000)} & \textcolor{black}{0.3816 (0.0000)} & \textcolor{black}{0.2438 (0.0000)} & \textcolor{black}{10.00}\\
				& \textcolor{black}{mPAC} & \textcolor{black}{0.3922 (0.0000)} & \textcolor{black}{0.5766 (0.0000)} & \textcolor{black}{0.5024 (0.0000)} & \textcolor{black}{0.6483 (0.0000)} & \textcolor{black}{9.00} \\
				& \textcolor{black}{COMIC} & \textcolor{black}{0.5142 (0.0688)} & \textcolor{black}{0.6085 (0.0672)} & \textcolor{black}{0.5319 (0.0846)} & \textcolor{black}{0.7433 (0.0616)} & \textcolor{black}{7.75}\\
				\cmidrule{2-7}
				& ${\text{MSCNL}}{{\text{G}}_{{\text{1-st}}}}$ & 0.6251 (0.0163) & 0.8423 (0.0170) & 0.7275 (0.0189) & 0.8731 (0.0137) & 3.00 \\
				& MSCNLG      & \bf{0.6989 (0.0011)} & \bf{0.8832 (0.0008)} & \bf{0.7913 (0.0012)} & \bf{0.9061 (0.0007)} & 1.00\\
				\midrule
				\multirow{10}{*}{Caltech101-20}  & AMGL        & 0.5224 (0.0576) & 0.4988 (0.0441) & 0.3837 (0.0400) & 0.7245 (0.0808) & 6.25\\
				& LMSC        & 0.6144 (0.0087) & 0.4729 (0.0295) & 0.3991 (0.0326) & 0.8641 (0.0052) & 4.75 \\
				& MLRSSC      & 0.5746 (0.0128) & 0.4174 (0.0167) & 0.3564 (0.0146) & 0.8601 (0.0027) & 7.50 \\
				& DMSCN       & 0.6249 (0.0007) & 0.4447 (0.0025) & 0.5067 (0.0001) & 0.8637 (0.0001) & 5.00\\
				& GMC         & 0.4809 (0.0000) & 0.4564 (0.0000) & 0.3403 (0.0000) & 0.5756 (0.0000) & 9.00 \\
				& \textcolor{black}{PMSC} & \textcolor{black}{0.4598 (0.0206)} & \textcolor{black}{\bf{0.5883 (0.0097)}} & \textcolor{black}{\bf{0.5622 (0.0373)}} & \textcolor{black}{0.8230 (0.0351)} & \textcolor{black}{4.50}\\
				& \textcolor{black}{mPAC} & \textcolor{black}{0.5135 (0.0000)} & \textcolor{black}{0.4493 (0.0000)} & \textcolor{black}{0.3484 (0.0000)} & \textcolor{black}{0.8218 (0.0000)} & \textcolor{black}{8.00}\\
				& \textcolor{black}{COMIC } & \textcolor{black}{0.6312 (0.0119)} & \textcolor{black}{0.4729 (0.0246)} & \textcolor{black}{0.4482 (0.0282)} & \textcolor{black}{0.8711 (0.0051)} & \textcolor{black}{3.50} \\
				\cmidrule{2-7}
				& ${\text{MSCNL}}{{\text{G}}_{{\text{1-st}}}}$ & 0.6333 (0.0006) & 0.4359 (0.0038) & 0.5153 (0.0004) & 0.8629 (0.0001) & 4.75 \\
				& MSCNLG      & \bf{0.6669 (0.0010)} & 0.4983 (0.0054) & 0.5393 (0.0068) & \bf{0.8763 (0.0014)} & 1.75 \\
				\midrule
				\multirow{10}{*}{LandUse21}     & AMGL        & 0.2772 (0.0148) & 0.1733 (0.0117) & 0.1126 (0.0055) & 0.6050 (0.0669) & 9.00\\
				& LMSC        & 0.3434 (0.0131) & 0.2967 (0.0127) & 0.2021 (0.0090) & 0.9192 (0.0026) & 4.00\\
				& MLRSSC      & 0.3188 (0.0046) & 0.2898 (0.0050) & 0.1828 (0.0024) & 0.9199 (0.0005) & 5.00\\
				& DMSCN       & 0.3691 (0.0022) & 0.2924 (0.0011) & 0.2340 (0.0012) & 0.9223 (0.0005) & 3.25 \\
				& GMC         & 0.2845 (0.0000) & 0.1371 (0.0000) & 0.1074 (0.0000) & 0.3954 (0.0000) & 9.50\\
				& \textcolor{black}{PMSC} & \textcolor{black}{0.2271 (0.0099)} & \textcolor{black}{0.2316 (0.0071)} & \textcolor{black}{0.1482 (0.0034)} & \textcolor{black}{0.8058 (0.0082)} & \textcolor{black}{8.25}\\
				& \textcolor{black}{mPAC} & \textcolor{black}{0.2863 (0.0000)} & \textcolor{black}{0.2805 (0.0000)} & \textcolor{black}{0.1810 (0.0000)} & \textcolor{black}{0.9127 (0.0000)} & \textcolor{black}{6.00} \\
				& \textcolor{black}{COMIC }      & \textcolor{black}{0.2938 (0.0082)} & \textcolor{black}{0.2195 (0.0114)} & \textcolor{black}{0.1491 (0.0056)} & \textcolor{black}{0.8922 (0.0050)} & \textcolor{black}{7.00} \\
				\cmidrule{2-7}
				&${\text{MSCNL}}{{\text{G}}_{{\text{1-st}}}}$ & 0.3898 (0.0105) & \bf{0.3443 (0.0128)} & 0.2545 (0.0115) & \bf{0.9274 (0.0011)} & 1.50\\
				& MSCNLG      & \bf{0.4028 (0.0039)} & 0.3362 (0.0074) & \bf{0.2631 (0.0082)} & 0.9265 (0.0004) & 1.50 \\
				\bottomrule
			\end{tabular}
		}
		\label{Comparision_Expreimental_Results1}
	\end{table}

	To demonstrate the effectiveness and superiority of the proposed MSCNLG, five state-of-the-art multi-view clustering methods are used for comparison, including AMGL \cite{AMGL}, LMSC \cite{LMSC}, MLRSSC \cite{MLRSSC_PR2018}, DMSCN \cite{DMSCN}, GMC \cite{GMC}, {\color{black}{PMSC \cite{2020PartitionKangZhao}, mPAC \cite{kang2019multipleIJCAI}, and COMIC \cite{peng2019comic}}}. Specifically, For LMSC, the dimension of latent representation is fixed to 100, and the optimal $\lambda$ is selected from $\{ 0.01,0.1,1,10,100\} $. For MLRSSC, parameters are referred to the recommended setting. For the DMSCN, autoencoder networks with six layers are employed, $\lambda_1$ and $\lambda_2$ are fixed to 1 and $1 \times {10^{\frac{c}{{10}} - 3}}$ respectively as suggested. {\color{black}{For PMSC and mPAC, we use the parameters as recommended in original papers.}} For AMGL and GMC, multi-view clustering results can be achieved without parameter selection. {\color{black}{Regarding to COMIC, we use k-means algorithm to help COMIC get exact clusters.}} Additionally, to further illustrate the effectiveness of the proposed method, the following baseline is also defined for comparison: ${\text{MSCNL}}{{\text{G}}_{{\text{1-st}}} }$, the fused multi-view graph of which is obtained only by the first-order proximity only. For the fair comparison, autoencoder networks of ${\text{MSCNL}}{{\text{G}}_{{\text{1-st}}} }$ and MSCNLG have six layers, the optimal values of $\alpha$ and $\beta$ are fixed to 0.1 and 100 respectively, 20 nearest neighbors are employed for both the first-order and second-order graph construction.
	
	\begin{table*}[t]
		\centering
		\resizebox{1\textwidth}{!}{
			\begin{tabular}{l|l|c|c|c|c|c|c|c|c}
				\toprule
				Dataset & View & \multicolumn{2}{c|}{NMI} & \multicolumn{2}{c|}{ACC} & \multicolumn{2}{c|}{F-measure} & \multicolumn{2}{c}{RI} \\ 
				\midrule
				\multirow{3}{*}{Yale Face}
				& View 1: ${\bf{X}}^{(1)} {\kern 3pt}| {\kern 3pt}{\bf{Z}}^{(1)}$ & \bf{0.5948} & 0.5584 & \bf{0.5733} & 0.5133 & \bf{0.4103} & 0.3569 & \bf{0.9263} & 0.9191\\
				& View 2: ${\bf{X}}^{(2)} {\kern 3pt}| {\kern 3pt}{\bf{Z}}^{(2)}$ & 0.2483 & \bf{0.6376} & 0.2075 & \bf{0.6010} & 0.0695 & \bf{0.4479} & 0.8736 & \bf{0.9310}\\
				& View 3: ${\bf{X}}^{(3)} {\kern 3pt}| {\kern 3pt}{\bf{Z}}^{(3)}$ & 0.2543 & \bf{0.6508} & 0.2085 & \bf{0.6034} & 0.0674 & \bf{0.4712} & 0.8802 & \bf{0.9343}\\
				\midrule
				\multirow{3}{*}{ORL}
				& View 1: ${\bf{X}}^{(1)} {\kern 3pt}| {\kern 3pt}{\bf{Z}}^{(1)}$ & \bf{0.7903} & 0.6909 & \bf{0.6275} & 0.4753 & \bf{0.5064} & 0.3282 & \bf{0.9765} & 0.9682\\
				& View 2: ${\bf{X}}^{(2)} {\kern 3pt}| {\kern 3pt}{\bf{Z}}^{(2)}$ & 0.4810 & \bf{0.7484} & 0.2450 & \bf{0.5758} & 0.0869 & \bf{0.4388} & 0.9545 & \bf{0.9735}\\
				& View 3: ${\bf{X}}^{(3)} {\kern 3pt}| {\kern 3pt}{\bf{Z}}^{(3)}$ & 0.4366 & \bf{0.7579} & 0.2017 & \bf{0.5741} & 0.0536 & \bf{0.4439} & 0.9522 & \bf{0.9737}\\
				\midrule
				\multirow{6}{*}{MSRCV1}
				& View 1: ${\bf{X}}^{(1)} {\kern 3pt}| {\kern 3pt}{\bf{Z}}^{(1)}$ & 0.3768 & \bf{0.4857} & 0.4952 & \bf{0.5460} & 0.3694 & \bf{0.4467} & 0.8221 & \bf{0.8408}\\
				& View 2: ${\bf{X}}^{(2)} {\kern 3pt}| {\kern 3pt}{\bf{Z}}^{(2)}$ & 0.0839 & \bf{0.3667} & 0.2381 & \bf{0.4810} & 0.1632 & \bf{0.3364} & 0.7537 & \bf{0.8136}\\
				& View 3: ${\bf{X}}^{(3)} {\kern 3pt}| {\kern 3pt}{\bf{Z}}^{(3)}$ & 0.2601 & \bf{0.6182} & 0.3524 & \bf{0.6714} & 0.2714 & \bf{0.5935} & 0.7766 & \bf{0.8854}\\
				& View 4: ${\bf{X}}^{(4)} {\kern 3pt}| {\kern 3pt}{\bf{Z}}^{(4)}$ & 0.2655 & \bf{0.5889} & 0.4238 & \bf{0.6873} & 0.2782 & \bf{0.5640} & 0.7897 & \bf{0.8772}\\
				& View 5: ${\bf{X}}^{(5)} {\kern 3pt}| {\kern 3pt}{\bf{Z}}^{(5)}$ & 0.0623 & \bf{0.4822} & 0.1667 & \bf{0.5492} & 0.2410 & \bf{0.4338} & 0.1994 & \bf{0.8334}\\
				& View 6: ${\bf{X}}^{(6)} {\kern 3pt}| {\kern 3pt}{\bf{Z}}^{(6)}$ & 0.1549 & \bf{0.1598} & 0.2905 & \bf{0.3111} & 0.2066 & \bf{0.2041} & 0.7750 & \bf{0.7757}\\
				\midrule
				\multirow{3}{*}{\textcolor{black}{LandUse-21}}
				& \textcolor{black}{View 1: ${\bf{X}}^{(1)} {\kern 3pt}| {\kern 3pt}{\bf{Z}}^{(1)}$} & \textcolor{black}{0.2037} & \textcolor{black}{\bf{0.2449}} & \textcolor{black}{0.1929} & \textcolor{black}{\bf{0.2257}} & \textcolor{black}{0.1069} & \textcolor{black}{\bf{0.1283}} & \textcolor{black}{0.9140} & \textcolor{black}{\bf{0.9160}}\\
				& \textcolor{black}{View 2: ${\bf{X}}^{(2)} {\kern 3pt}| {\kern 3pt}{\bf{Z}}^{(2)}$} & \textcolor{black}{0.2183} & \textcolor{black}{\bf{0.2556}} & \textcolor{black}{0.1971} & \textcolor{black}{\bf{0.2062}} & \textcolor{black}{0.1179} & \textcolor{black}{\bf{0.1243}} & \textcolor{black}{0.9132} & \textcolor{black}{\bf{0.9162}}\\
				& \textcolor{black}{View 3: ${\bf{X}}^{(3)} {\kern 3pt}| {\kern 3pt}{\bf{Z}}^{(3)}$} & \textcolor{black}{0.2122} & \textcolor{black}{\bf{0.2246}} & \textcolor{black}{0.1905} & \textcolor{black}{\bf{0.1943}} & \textcolor{black}{0.1080} & \textcolor{black}{\bf{0.1145}} & \textcolor{black}{0.9076} & \textcolor{black}{\bf{0.9118}}\\
				\bottomrule
			\end{tabular}
		}
		\caption{Subspace clustering results performed on original data representations ${\bf{X}}^{(v)}$ and latent representations ${\bf{Z}}^{(v)}$ via the low-rank representation. \textcolor{black}{Experimental results on Yale Face, ORL, MSRCV1, and LandUse-21 are reported here.} For the fair comparison, LRR with a fixed tradeoff parameter ($\lambda=0.01$) is employed for all datasets and all views. Values in the bold type denote the best clustering.}\smallskip
		\label{Analysis_Expreiments}
	\end{table*}
	
	\subsection{Comparison Experiments}
	As displayed in Table \ref{Comparision_Expreimental_Results} and \ref{Comparision_Expreimental_Results1}, clustering results in metrics of NMI, ACC, F-measure, and RI with corresponding ranks are reported. In a big picture, the proposed deep multi-view smooth representation clustering network obtains promising clustering results, and outperforms other comparison approaches with a large margin. And comparison experimental results illustrate the following observations: 
	\begin{itemize}
		\item [1)]  Compared with traditionally multi-view subspace clustering methods, the proposed MSCNLG achieves significant improvements. For example, on MSRCV1 dataset, 20.15\% and 6.41\% increments are attained in metrics of NMI and RI respectively with respect to MLRSSC. Since autoencoder networks used in the proposed method can reconstruct data of all views, and reconstructed data can be more suitable for the linear subspace model, it is expected that the proposed method can achieve better clustering results than these traditional approaches. 
		\item [2)]  The proposed method also gains remarkable progress over another deep-based algorithm, i.e., DMSCN, owing to the underlying local and global multi-view graph information are used in our method. 
		\item [3)]  Compared with MSCNL$\text{G}_\text{1-st}$, the proposed MSCNLG investigates multi-view graph information by the first-order and second-order proximity, and it achieves better clustering results than MSCNL$\text{G}_\text{1-st}$ in most cases, which means that the local and global multi-view graph information plays an important role in our method.
	\end{itemize}
	
	Are Autoencoder Networks Necessary? The autoencoder networks employed in multiple views can reconstruct the input data, that is to say that data attained from the output of encoder networks are more suitable for the linear subspace model. Specifically, we conduct a low-rank subspace clustering algorithm \cite{LRR_PAMI2012} on original data of multiple views and latent representations obtained from the output of encoder networks to illustrate the rationality and necessity of leveraging autoencoder networks. 
	
	Experiments on Yale Face, ORL, MSRCV1, and LandUse-21 are taken for examples. As can be observed in Table~\ref{Analysis_Expreiments}, clustering results demonstrate that the reconstructed data ${\bf{Z}}^{(k)}$ are much suitable for subspace clustering than original input data ${\bf{X}}^{(k)}$ in general. {\color{black}{It is noteworthy that clustering results of the first view's latent representations of Yale Face and ORL are slightly worse, the reason may be that during the learning process, the first view's latent representations of these two datasets are influenced by the rest two views. However, it can be observed that clustering results of latent representations gain remarkable improvements in most cases. Therefore, it is necessary to employ the autoencoder networks in the proposed MSCNLG to reconstruct the multi-view data for clustering.}}
	
	\subsection{Model Analysis}
	Taking the experiments on Yale Face for example, the running time,  parameter sensitivity, and convergence property are also discussed in this section to further verify the effectiveness of our method. 
	
	{\color{black}{
			\subsubsection{Running Time}
			We conduct comparison methods on the desktop with GeForce
			GTX 1080, Inter Core i7 4-core process with 3.60GHz, and 8GB RAM. The running time of different methods on different datasets are reported in Table~\ref{Running_Time}. It can be observed that the running time of the proposed MSCNLG is medium among comparison methods.	}}
	\begin{table}[!htbp]
		\centering
		\resizebox{0.85\textwidth}{!}{
			\begin{tabular}{l|c|c|c|c|c|c}
				\toprule
				\textcolor{black}{Method} & \textcolor{black}{Yale Face} & \textcolor{black}{ORL} & \textcolor{black}{MSRCV1} & \textcolor{black}{BBC} & \textcolor{black}{Caltech101-20} & \textcolor{black}{LandUse-21} \\ 
				\midrule
				\textcolor{black}{AMGL} & \textcolor{black}{1.0179} & \textcolor{black}{1.0247} & \textcolor{black}{0.2703} & \textcolor{black}{2.5898} & \textcolor{black}{55.6623} & \textcolor{black}{61.3051} \\  
				\textcolor{black}{LMSC} & \textcolor{black}{21.0671} & \textcolor{black}{43.0265} & \textcolor{black}{6.6855} & \textcolor{black}{111.1109} & \textcolor{black}{1207.0062} & \textcolor{black}{770.7249} \\  
				\textcolor{black}{MLRSSC} & \textcolor{black}{0.8532} & \textcolor{black}{0.7055} & \textcolor{black}{0.1899} & \textcolor{black}{4.2883} & \textcolor{black}{45.4016} & \textcolor{black}{17.3046} \\
				\textcolor{black}{DMSCN} & \textcolor{black}{13.1850} & \textcolor{black}{13.6113} & \textcolor{black}{15.3801} & \textcolor{black}{37.0080} & \textcolor{black}{65.1870} & \textcolor{black}{30.8490} \\
				\textcolor{black}{GMC} & \textcolor{black}{1.7270} & \textcolor{black}{1.3814} & \textcolor{black}{0.7698} & \textcolor{black}{57.7365} & \textcolor{black}{36.4361} & \textcolor{black}{18.9073} \\
				\textcolor{black}{PMSC} & \textcolor{black}{18.8195} & \textcolor{black}{76.2700} & \textcolor{black}{7.1981} & \textcolor{black}{112.3667} & \textcolor{black}{910.9132} & \textcolor{black}{321.2634} \\
				\textcolor{black}{mPAC} & \textcolor{black}{1.7505} & \textcolor{black}{6.8296} & \textcolor{black}{4.2687} & \textcolor{black}{107.2394} & \textcolor{black}{1400.9071} & \textcolor{black}{535.6096} \\
				\textcolor{black}{COMIC} & \textcolor{black}{8.3820} & \textcolor{black}{26.4751} & \textcolor{black}{1.6682} & \textcolor{black}{75.6783} & \textcolor{black}{99.3251} & \textcolor{black}{11.3584} \\
				\midrule
				\textcolor{black}{MSCNLG$_{1\rm{st}}$} & \textcolor{black}{12.7250} & \textcolor{black}{20.7930} & \textcolor{black}{17.8160} & \textcolor{black}{41.5864} & \textcolor{black}{1615.6146} & \textcolor{black}{152.1420} \\
				\textcolor{black}{MSCNLG} & \textcolor{black}{12.9950} & \textcolor{black}{21.0299} & \textcolor{black}{17.9400} & \textcolor{black}{42.0190} & \textcolor{black}{1772.3475} & \textcolor{black}{153.0880} \\ 
				\bottomrule
		\end{tabular}}
		\caption{\textcolor{black}{Running time of comparison methods on all datasets. The unit of time is seconds.}}
		\label{Running_Time}
	\end{table}

	\subsubsection{Parameter Sensitivity}
	Moreover, parameter sensitivity is also explored. In metrics of NMI and ACC, {\color{black}{we show clustering results by selecting values of $\alpha$ and $\beta$ from $\{$0.001, 0.01, 0.1, 1, 10, 100, 1000, 10000$\}$. As can be observed in Fig.~\ref{Results_Analysis}(a) and Fig.~\ref{Results_Analysis}(b), the proposed method is relatively insensitive to $\beta$, and can achieve promising results when $\alpha$ is less than 0.1.}} Regrading to the number of nearest neighbors in multi-view graph fusion, Fig.~\ref{Results_Analysis}(c) illustrates that the proposed approach is pretty robust to the number of nearest neighbors. Therefore, the proposed MSCNLG is robust for different parameters.
	
	\subsubsection{Convergence Property}
\begin{figure*}[t]
		\centering
		\includegraphics[width=1.0\textwidth]{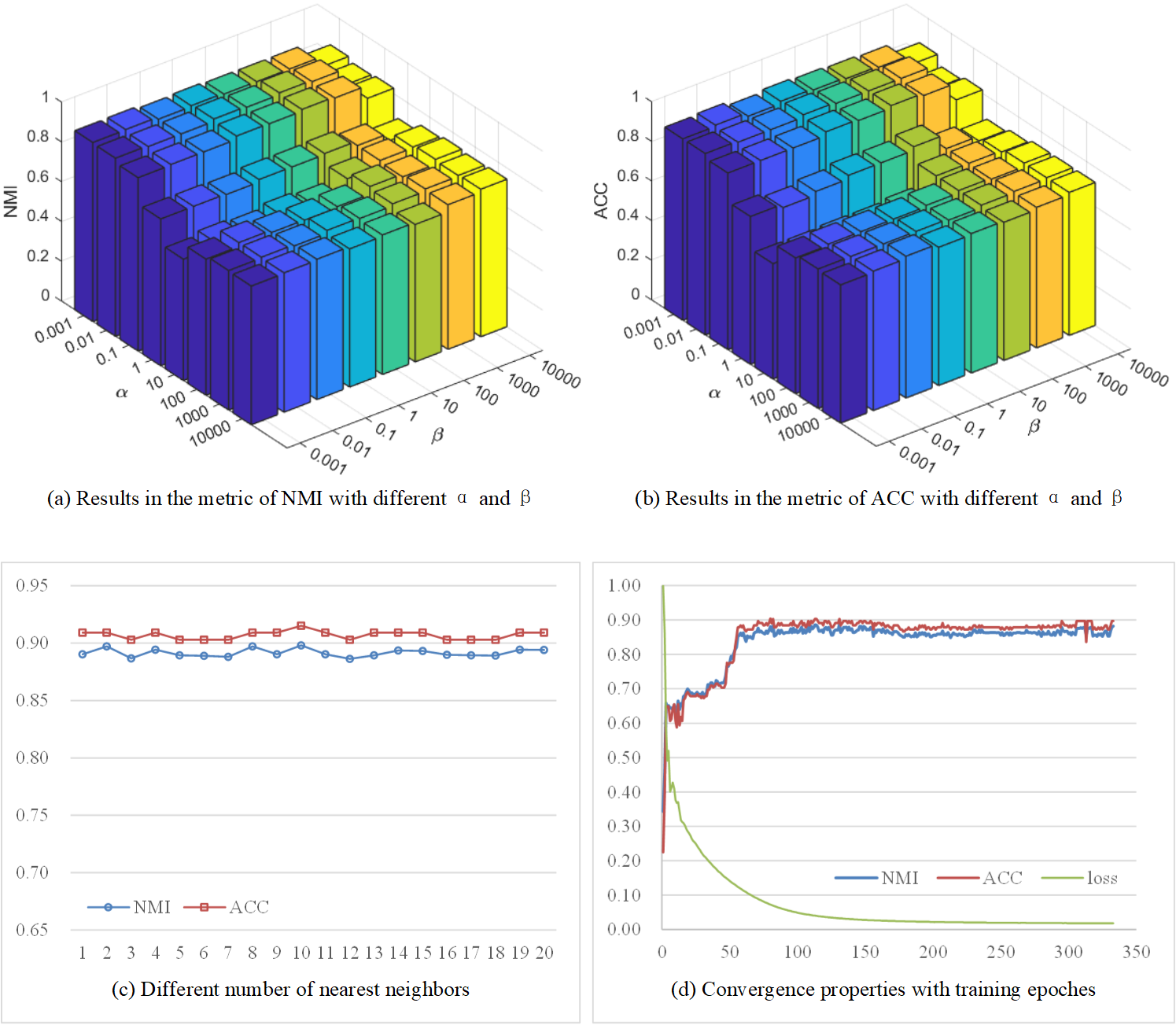} 
		\caption{Model analysis of the proposed method. {\color{black}{(a) and (b) present clustering results in metrics of NMI and ACC with respect to different $\alpha$ and $\beta$.}} (c) illustrates the influence of the nearest neighbors number in the multi-view graph fusion. (d) shows the convergence properties of the proposed MSCNLG. }
		\label{Results_Analysis}
	\end{figure*}

	Taking experiments conducted on Yale Face for example. As depicted in Fig.~\ref{Results_Analysis}(d), the normalized loss and clustering results (in metrics of NMI and ACC) with respect to epochs are reported here. Actually, experimental results on other datasets have the similar performance. It is clear that the proposed method has a stable convergence property.
	
	\section{Conclusion} 
	In this paper, the novel multi-view subspace clustering networks with local and global graph information is proposed to achieve promising multi-view clustering results. By leveraging the autoencoder and exploring the underlying local and global graph information, the desired subspace representation can be obtained in our method. Extensive experiments conducted on six benchmark datasets illustrate the effectiveness and competitiveness of the proposed method in comparison to several state-of-the-art multi-view clustering methods.
	
	\section*{Acknowledgements}
	
	This work is supported by the Fundamental Research Funds for Central Universities under Grant No. xzy022020050.

	\bibliography{MSCNLG}
	
\end{document}